\documentclass{soups}

\usepackage{balance}       % to better equalize the last page
\usepackage{graphics}      % for EPS, load graphicx instead 
\usepackage[T1]{fontenc}   % for umlauts and other diaeresis
\usepackage{txfonts}
\usepackage{mathptmx}
\usepackage[pdflang={en-US},pdftex]{hyperref}
\usepackage{color}
\usepackage{booktabs}
\usepackage{textcomp}

\usepackage{microtype}        % Improved Tracking and Kerning
\usepackage{ccicons}
\usepackage{todonotes}
\usepackage{algpseudocode,algorithm,algorithmicx}
\usepackage{tabularx}
\usepackage{amsmath}
\usepackage{xspace}
\usepackage{caption}
\usepackage{graphicx}
\usepackage{caption}
\usepackage{subfig}
\usepackage{multirow}
\usepackage{amsmath}
\usepackage{colortbl}
\usepackage{fixltx2e}
\usepackage{hyperref}
\usepackage[font={small,it}]{caption}
\usepackage{amssymb}

\def\plaintitle{Distinctiveness, Complexity, and Repeatability of Online Signature Templates\titlenote{Published as \cite{SAEBAE2018332}}}

\def\emptyauthor{}
\def\plainkeywords{Biometric; Online signature; Handwritten signature; Template characteristics; Biometric quality; Distinctiveness; Complexity; Repeatability}

\makeatletter
\def\url@leostyle{%
  \@ifundefined{selectfont}{
    \def\UrlFont{\sf}
  }{
    \def\UrlFont{\small\bf\ttfamily}
  }}
\makeatother
\def\pprw{8.5in}
\def\pprh{11in}

\definecolor{linkColor}{RGB}{6,125,233}
\hypersetup{%
  pdftitle={\plaintitle},
  pdfauthor={\emptyauthor},
  pdfkeywords={\plainkeywords},
  pdfdisplaydoctitle=true, % For Accessibility
  bookmarksnumbered,
  pdfstartview={FitH},
  colorlinks,
  citecolor=black,
  filecolor=black,
  linkcolor=black,
  urlcolor=linkColor,
  breaklinks=true,
  hypertexnames=false
}

\begin{document}

\title{\plaintitle}

\numberofauthors{2}
\author{%
 \alignauthor{Napa~Sae-Bae\\
    %\affaddr{for Submission}\\
    \affaddr{Rajamangala University of Technology Suvarnabhumi, Thailand}\\
    \email{napa.s@rmutsb.ac.th, benapa@gmail.com}}\\
 \alignauthor{Nasir Memon\\
    %\affaddr{for Submission}\\
    \affaddr{New York University, USA}\\
    \email{memon@nyu.edu}}\\
 \and    \alignauthor{Pitikhate~Sooraksa\\
    %\affaddr{for Submission}\\
    \affaddr{King Mongkut's Institute of Technology Ladkrabang, Thailand}\\
    \email{pitikhate.so@kmitl.ac.th}}\\
}

\maketitle

\begin{abstract}

This paper proposes three measures to quantify the characteristics of online signature templates in terms of distinctiveness, complexity and repeatability. A distinctiveness measure of a signature template is computed from a set of enrolled signature samples and a statistical assumption about random signatures. Secondly, a  complexity measure of the template is derived from a set of enrolled signature samples. Finally, given a signature template, a measure to quantify the repeatability of the online signature is derived from a validation set of samples. These three measures can then be used as an indicator for the performance of the system in rejecting random forgery samples and skilled forgery samples and the performance of users in providing accepted genuine samples, respectively. The effectiveness of these three measures and their applications are demonstrated through experiments performed on three online signature datasets and one keystroke dynamics dataset using different verification algorithms.
  
\end{abstract}


\section{Introduction}
For the past decade or so, research in biometric verification systems has mostly focused on effective feature sets and recognition algorithms to improve verification performance~\cite{parodi2014legendre,6786375,cpalka2014new,van2014finger, nguyen2017draw, NGUYEN2018174, nguyen2017smartwatches, van2018user}. However, several studies have indicated that characteristics of biometric samples used in the enrollment process to create a verification template can also play a crucial role in the overall accuracy and reliability of biometric systems~\cite{alonso2012quality,yager2010biometric,unar2014review}. Specifically, the accuracy of face, fingerprint, and iris biometric systems evaluated on good quality biometric samples is much higher than systems evaluated on poor quality samples~\cite{jain201650}.
In text password based authentication where users are allowed to choose a password freely, it is known that some users would choose a weak password or a password that could be guessed easily~\cite{komanduri2014telepathwords}. This problem has led to many studies related to assessing password strength and also the development of mechanisms such as the password strength meter that offers feedback and enforces a policy to reject a weak password during the enrollment stage itself~\cite{komanduri2014telepathwords}.%kelley2012guess

Similarly, there is a need for the assessment of biometric template characteristics when enrollment is done without supervision~\cite{labati2016biometric}, e.g., for mobile device authentication, and rejecting samples that could lead to degradation of performance.  This is especially true for behavioral biometric based authentication techniques such as online signatures or key stroke dynamics. In behavioral biometric verification systems it is difficult to collect a set of negative samples that could accurately represent a population. Therefore, a user specific template is generally modeled based purely on a small number of enrolled samples, without using imposter samples. Hence, at a given threshold, the verification performance can be profoundly different from one template to another.

\begin{figure*}
    \centering
    \subfloat[Random forgery]
    {\includegraphics[width=0.9\columnwidth]{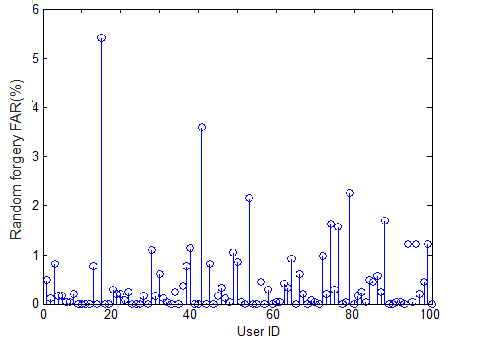}}\quad
    \subfloat[Skilled forgery]
    {\includegraphics[width=0.9\columnwidth]{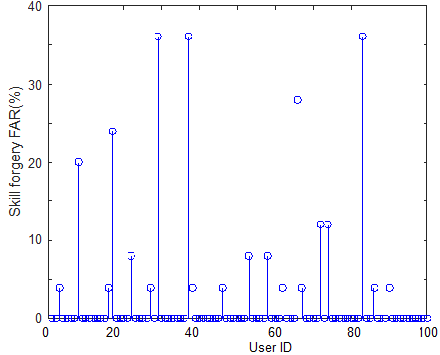}}
    \vspace{-0.5em}
    \caption{FAR against random forgery and skilled forgery for each  user in MCYT dataset when a user's signature template is derived from the first 10 genuine samples and threshold set at EER with respect to the algorithm described in~\protect \cite{6786375} %(the mean of FAR is at 0.42\% with standard deviation of 0.77)
    }
    \label{fig:FAR_RF}
\end{figure*}

For example, in online signature authentication, a user may enroll a template that is very simple. As a result, attackers could easily produce a signature that matches the template, resulting in a compromised account. In this context, an assessment of biometric template characteristics could be used to design proper mechanisms to cope with such weak templates~\cite{bharadwaj2015qfuse,huang2015adaptive,ross2009exploiting,vivaracho2016client}. For example, given a template that is predicted to yield high FAR (False Acceptance Rate: the likelihood that forgery samples will be incorrectly accepted by the system) but low FRR (False Rejection Rate: the likelihood that genuine samples will be incorrectly rejected by the system), the system could examine whether a few bad enrolled samples could be safely removed to lower FAR without degrading FRR. Or given a template that is predicted to yield high FRR, the system may prompt users to re-enroll or practice more with their signatures.

In this paper we focus on characteristics of one particular type of behavioral biometrics, namely online signatures. We develop three simple assessment characteristics of online signature templates and investigate their ability to predict FAR and FRR. 
The three characteristics are: {\bf 1) Distinctiveness} refers to how much this user's signature differs from others, which indicates the probability of a user being impersonated by a random signature, {\bf 2) Complexity} refers to how hard to forge this user's signature. This indicates the probability of a user being impersonated by a forged signature, and {\bf 3) Repeatability} refers to how likely this user can repeat her signature which indicates the probability of a genuine sample being rejected by the system.
%\end{itemize}

We show, using experiments,  how the assessment of these characteristics for online signature templates can be used to infer security and usability of a particular online signature template. Specifically, distinctiveness and complexity could be indicative of security or FAR of the signature template against random forgeries and skilled forgeries respectively. Also, repeatability could be indicative of usability in terms of FRR, which refers to the ability of the user in gaining access to the system using her signature.  Note that this work is an extension of previous work~\cite{sae2015quality} where we presented the metric to measure distinctiveness of online signature templates and preliminary results to demonstrate its efficacy.

\begin{figure}[!t]
    \centering
    \subfloat[The 25th percentile user]{\includegraphics[width=3.4in]{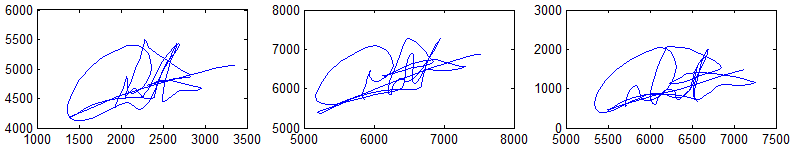}}\\
    \vspace{-0.5em}
    \subfloat[The 75th percentile user]{\includegraphics[width=3.4in]{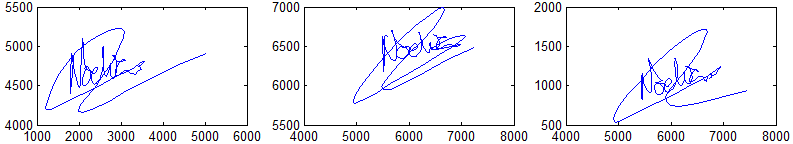}}
    \vspace{-0.5em}
    \caption{The 3rd, 7th, and 10th samples of two users from MCYT dataset with mean imposter scores at the 25th and 75th percentile}
    \label{fig:samples}
\end{figure}
\subsection{Performance disparity of signature templates} 
To illustrate the disparity in characteristics of online signature templates in terms of distinctiveness and complexity, FAR against random forgery and skilled forgery of each individual in MCYT dataset is plotted in Figure~\ref{fig:FAR_RF}. Note that EER denotes the equal error rate or the rate at which FAR and FRR are equal. 
Samples of enrolled signatures that lead to the mean imposter score at the 25th and 75th percentile are shown in Figure~\ref{fig:samples}. It is noticed that, in addition to the intrinsic difference of the two signatures, the first set of signatures (25th percentile) shows significant variation among enrolled samples. Consequently, the mean of the imposter score distribution of the first set is lower than the second. 
Note that the score here represents dissimilarity between a test sample and the signature template. A lower score means more similar to the template.

\begin{figure}
    \centering
%    \vspace{-1em}
    \subfloat[Simple signature]{\includegraphics[width=0.4\columnwidth]{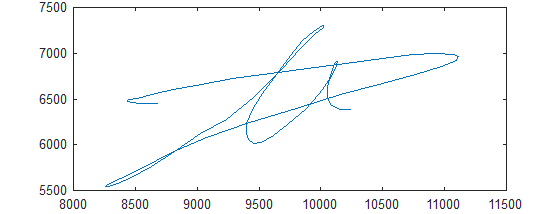}}\quad
    \subfloat[Complex signature]{\includegraphics[width=0.4\columnwidth]{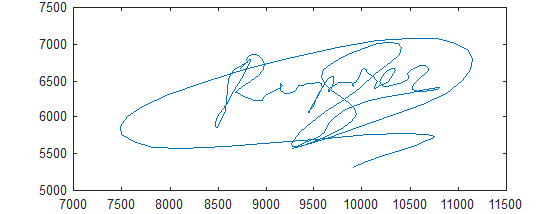}}
    \vspace{-0.5em}
    \caption{The samples of simple and complex signatures from MCYT dataset}%(the mean of FAR is at xxx\% with standard deviation of yyy)

    \label{fig:sample_complex}
\end{figure}

Similarly, the samples of signature that are considered simple and complex (the ones with higher and lower FAR respectively) are illustrated in Figure~\ref{fig:sample_complex}. That is, the more complex signature templates are the ones with the higher amount of directional and speed changes between two consecutive drawing vectors and also with consistent enrolled signature samples.

\begin{figure}
    \centering
        \includegraphics[width=\columnwidth]{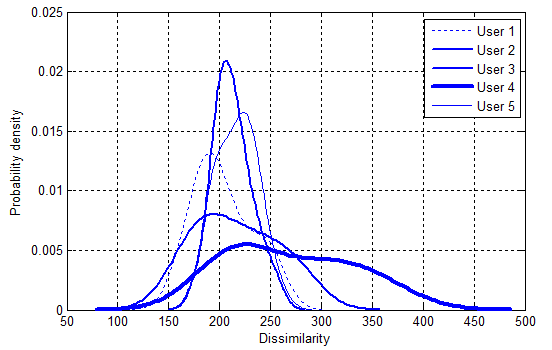}
        \vspace{-0.5em}
    \caption{Genuine score distribution of different users from NYU dataset with respected to the algorithm described in~\protect \cite{6786375}}
    \label{fig:score2}
\end{figure}

\begin{figure}[!t]
%\vspace{-0.5em}
    \centering
    \subfloat[The first session samples]{\includegraphics[width=\columnwidth]{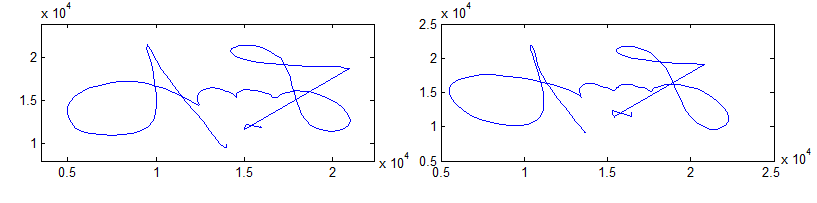}}\quad
    \subfloat[The second session samples]{\includegraphics[width=\columnwidth]{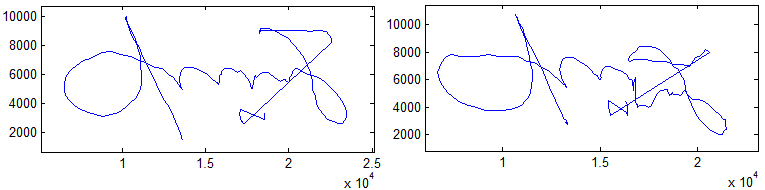}}
    \caption{The 4th and 8th samples of a user with high FRR from SUSIG dataset showing that the samples in the first session are consistent but they are not consistent with the ones from the second session}
    \label{fig:samples_repeatability}
\end{figure}

In addition to FAR, it is also known that in a biometric verification system, user contribution toward FRR is disproportional~\cite{rattani2012analysis}. This can also be observed in online signatures as illustrated in Figure~\ref{fig:score2}. %As can be seen  genuine attempts of some users will be rejected more often than others.
Repeatability of a biometric trait, i.e., similarity between enrolled signatures and genuine test samples with respect to a given recognition system, is different from user to user, as demonstrated in Figure~\ref{fig:samples_repeatability}.
That is, the templates constructed from enrolled samples that do not sufficiently capture variation among genuine samples would exhibit higher FRR. This in turn affects usability. In fact, repeatability of biometric features has been studied in many biometric modalities including ocular biometric~\cite{ferrer2017evaluation}, brain EEG~\cite{ruiz2017permanence}, and online signatures~\cite{guest2004repeatability}. These studies focused on evaluating repeatability of biometric features using samples across sessions in order to derive an optimal feature set to be used in the recognition system. However, they do not help in inferring user or template specific FRR. 

The rest of this paper is organized as follows. Related work on biometric characteristics or quality is discussed in Section~\ref{section:RelatedWork}. Next, the proposed method to assess three online signature template characteristics is described in Section~\ref{section:quality}. In Section~\ref{section:Experimental}, experimental results that demonstrate the effectiveness of the proposed metrics are presented. Conclusions are summarized in Section~\ref{section:Discussion}.

\section{Related work}
\label{section:RelatedWork}
The phenomenon of a ``biometric menagerie'', or ``biometric zoo'', has been observed in many biometric modalities and across several datasets~\cite{yager2010biometric,houmani2016hunting}. These terms refer to the observation that the contribution of each user towards the overall FAR and FRR of a biometric system can vary significantly. For example, lambs are users who have high match scores with random imposters thereby having a higher FAR. As a result of this observation, studies on the quality of biometric samples emerged in order to improve system performance and reliability. For instance, based on a quality measure, one could decide whether to: 1) alter the enrollment process including requesting users to re-enroll in the system, 2) adjust the decision threshold~\cite{vivaracho2016client} or weighting function in a multimodal biometric system, 3) invoke different preprocessing or recognition algorithms~\cite{bharadwaj2015qfuse}, or 4) update the template~\cite{grother2007performance}. A quality score could also be used as a deciding factor whether an additional biometric trait has to be required~\cite{ross2009exploiting}.

Automated quality measurement of biometric samples has been explored in many modalities, e.g., fingerprint, iris, and speaker recognition. One typical approach is to measure the quality of the signal extracted from a biometric sensor. For example, in face and ear biometric, the sparse coding error which can detect expression, pose and random pixel corruption has been proposed as a quality metric~\cite{huang2015adaptive}. In speaker verification, signal-to-noise ratio (SNR) and the perceived quality of speech utterance have been used~\cite{villalba2016analysis}. This approach is applicable to physiological biometrics where signal quality of a biometric sample plays an important role in recognition performance for a particular user. %and it is affected by background environment

However, the general concept of behavioral biometric recognition system is quite different from that of a physiological one. In a behavioral biometric system, a user's biometric sample is collected over a period of time and the user's template or a description of a user-specific classifier is typically derived from characteristics of the enrolled samples as well as their intra-user variation. Furthermore, for some types of behavior biometrics, there is user choice in creating the template. For example with online signatures, a user can select from multiple signatures. Therefore, the characteristics of the behavioral biometric templates which infer recognition performance of the system for a particular user, depends on multiple factors including the character or essence of an individual biometric trait, the distinctiveness of the underlying content of the template, the consistency between the samples used for enrollment and the repeatability between the enrolled samples and test samples~\cite{cho2006artificial}.

One approach that is more suitable for behavioral biometrics is to quantify the amount of biometric information embedded in biometric samples. For example, Kullback-Leibler divergence between a feature vector of a query sample and the average feature vector from all enrollment samples has been proposed as the quality metric in keystroke dynamic~\cite{morales2014towards}.
Regarding handwritten signature modalities, Zois et al.,~\cite{zois2016offline} has proposed overall quality characterization of genuine signatures by means of stability and complexity derived from distribution statistics and stroke characteristics. For the specific case of online signature systems, a research attempt to develop a quality metric for a set of enrolled samples based on local density estimation by a hidden markov model has been made~\cite{houmani2014quality}. In addition, a quality metric for online signatures based on a sigma-lognormal model which measures kinematic information exhibited in online signature samples has been developed~\cite{galbally2011quality}. These studies, along with the one in~\cite{houmani2012biosecure}, have demonstrated that, enrolled signatures with different quality ranges result in different verification performance. However, it is not clear that the observed performance discrepancy is caused by the difference in genuine or imposter score distributions. In other words, the type of a user in the biometric menagerie can not be identified using these quality metrics. As a result, it is difficult to refine selection strategies to deal with problematic templates~\cite{poh2008incorporating} thereby limiting the value of the quality metric. In addition, the concept of a user's personal entropy and relative entropy to infer a signature template's FRR and FAR against skilled forgery has recently been introduced by Houmani et al., in 2016~\cite{houmani2016hunting}. In their work, the proposed user's relative entropy for each signature template is derived from Kullback-Leibler distance between the local probability density functions of HMM model of genuine samples and that of skilled forgeries. The drawback of this approach is the need for skilled forgery samples, which are not generally available.

In contrast to the above, this paper focuses on assessing three characteristics relevant to the quality of online signature templates, namely distinctiveness, complexity, and repeatability scores using only genuine samples. These metrics can then be used as an indicator for imposter score distributions of random forgeries and skilled forgery as well as genuine score distributions, separately. The characteristics proposed in this paper have multiple benefits. First, the distinctiveness and complexity of an online signature template is directly related to the likelihood of a user being a lamb, i.e., a user who is vulnerable to impersonation~\cite{yager2010biometric}. The distinctiveness and complexity metric can be used to detect such users and depending on the application, require them to enroll with a new signature or to provide more samples of the signature to substitute  inconsistent ones. In addition, the repeatability metric can be used to detect goat users, i.e. the users with signatures that have low repeatability thereby resulting in high false rejection rates~\cite{yager2010biometric, poh2013user}. In this case, more training sessions might be required for a user's signature to stabilize. These mechanisms could be particularly useful when the biometric system operates without supervision as in the case of user authentication on mobile devices.  

\section{Proposed Online signature template characteristics}
\label{section:quality}
In this section, the metrics to assess distinctiveness, complexity, and repeatability of online signature templates are presented. Specifically, distinctiveness and complexity are computed based on a subset of histogram features from a recently proposed histogram-based verification system~\cite{6786375}. That is, given an online signature instance represented by a time-series of data points where the attributes of each data point include x-y coordinates, pressure, etc., the histograms for this signature sample are then computed from time-series of the attributes and their derivatives~\cite{6786375}.
Repeatability is computed based on genuine scores of enrolled samples and validation samples. %Details for each metric are given below.
\subsection{Template distinctiveness}
\label{subsection:distinctiveness}
Since many verification algorithms do not assume existence of forgery samples, online signature verification algorithms typically normalize sample attributes, e.g., similarity/dissimilarity score, features, etc., by the distribution of training samples~\cite{kholmatov2009susig,6786375,manjunatha2016online}. 
Consequently, when a system is operated at the same decision threshold level for all users, signatures with different characteristics will correspond to different levels of FAR. In other words, some users would have less protection against random forgery than others. This subsection describes a method to derive a signature template distinctiveness score that can be indicative of such problematic templates. The proposed method is based on statistical discrimination of features derived from a set of genuine online signatures and a generic assumption about random signatures. Specifically, the features used here are histograms of speed-angle and pressure derived from the first and second halves of an online signature.

The first step is to compute statistics of feature distributions of user signature samples and random signature samples. The statistics of feature distributions of user signature samples can be derived directly from the set of enrolled signature samples. However, it is not a trivial task to collect a set of random forgery samples. In addition, statistics derived from actual signatures in a dataset could depend on the signing language, acquisition device etc., and could vary across datasets. Therefore, in order to compute statistics of random forgery samples, a general assumption about random signatures is made specifically about the features used to compute this metric rather than using a set of random signatures. This is in fact a great benefit of deriving statistics in feature spaces instead of using score distribution where a collection of random signature samples is required. Then the second step is to compute a decidability index~\cite{daugman2003importance}, which is a measure of separation between feature distribution of user signature samples and random signature samples, for each of the features. Lastly, the distinctiveness score is computed from a sum of each feature decidability index.

Let $(\mu^T_{f_i},\sigma^T_{f_i})$ and $(\mu^P_{f_i},\sigma^P_{f_i})$ be the mean and standard deviation of the $i^{th}$ feature distributions ${f_i}$ of user-signature samples and random signature samples.  The decidability index for  the $i^{th}$  feature, denoted by $d_{f_i}$, for these two distributions is defined as follows:
\begin{equation}
d_{f_i} = \frac{\|\mu^T_{f_i}-\mu^P_{f_i}\|}{\sqrt{(\sigma^T_{f_i}+\sigma^P_{f_i})/2}}
\end{equation}

%Then, given a biometric template with $M$ features,
For $\mu^T_{f_i}$ and $\sigma^T_{f_i}$, they can be derived from the feature distribution of the user's enrolled samples in the template. 

Then, the problem of deriving feature distribution statistics  $(\mu^P_{f_i},\sigma^P_{f_i})$ of random forgery samples is addressed by using global statistics, instead of dataset specific ones. That is, since the feature set that is used to compute the distinctiveness score of signature templates comprises of frequency of histogram bins from speed-angle and pressure histograms, the statistics of the distribution of each histogram bin frequency can be estimated using a generic assumption about random signatures as follows.

Let $R = \{r_i | i = 1, 2,..., L\}$ be the time series of a particular attribute from a random signature of length $L$  that is used to derive a particular histogram. Let also assume that the probability of $r_i$ falling into any histogram bin be uniform and independent. That is, the probability that ${r_i}$ falls into any bin of histogram equals to $\frac{1}{N}$ %and not falling into the bin equals to $1-\frac{1}{N}$
where $N$ is the total number of bins that appear in the given histogram.
Consequently, the probability distribution %of each histogram attribute, or
of the number of elements $r_i$ falling into a particular bin follows a binomial distribution. That is, the probability that a histogram bin has a value $k$ is given by:
\begin{equation}
P(k) = \binom{L}{k} p^k(1-p)^{L-k}
\label{eq:normal}
\end{equation}
where  $P(k)$ is the probability that $k$ elements (from the total of $L$ elements) fall into a particular bin of the histogram and $p=\frac{1}{N}$ is the probability that ${r_i}$ falls into a particular bin of an $N$-bin histogram. 

As a result, the mean ($\mu$) and standard deviation ($\sigma$) of each histogram bin value according to the distribution in~(\ref{eq:normal}) can be derived as:
\begin{equation}
%\begin{split}
\mu^P_{f_i} = L\times p = L\times\frac{1}{N} \text{ ,}\\
\end{equation}
\begin{equation}
\text{and}\quad \sigma^P_{f_i} = \sqrt{L \times p\times(1-p)} = \sqrt{L \times \frac{(N-1)}{N^2}}
\end{equation}

The above gives the statistics of the histogram feature distribution of random signatures when a histogram is attributed by the actual count of elements falling into a particular histogram bin. And for a relative frequency histogram, the mean and standard deviation of the histogram feature is further divided by $L$. That is:
\begin{equation}
\mu^P_{f_i} = \frac{1}{N} \text{ ,}\\%121.8670-susig, 215.3070--mcyt
% 351.5660--mcyt  237.6500--susig
\end{equation}
\begin{equation}
\text{and}\quad \sigma^P_{f_i} = \sqrt{\frac{1}{L} \times \frac{(N-1)}{N^2}}
\end{equation}

Then, the overall distinctiveness $D(T)$ of the template $T$  is defined by:
\begin{equation}
D(T) = \sum_{=1}^{M}{d^T_{f_i}}
\end{equation}
where $M$ is the total number of features used to compute the distinctiveness score. Note that the histogram features used here are the ones from speed-angle and pressure histograms derived from the first and second
halves of an online signature.

This template distinctiveness score $D(T)$ takes both the uniqueness and consistency of enrolled signature samples in the template under consideration. In particular, the difference of feature mean values between random forgery samples and enrolled signature samples of a user could be viewed as uniqueness of that user's signature. Similarly, feature variance values derived from the enrolled samples of each user could be viewed as the inverse of consistency of the user's enrolled signature samples.

\subsection{Template complexity}
\label{subsection:complexity}
The complexity score of an online signature template is a security measure against skilled forgery attempts. That is, the more complex signature templates are the ones that are harder to forge.%Using the same set of features, the complexity of online signature template can be computed as follows.
The proposed signature complexity score is computed using the histogram features based on two factors: the degree of signature complexity and the dispersion index function of a signature. %Details of these two functions are given below.
\subsubsection{Degree of signature complexity:} Intuitively, the simplest form of a signature is to continuously draw in the same manner, i.e.,  one where a signature has no change in the drawing angle or speed. Given a histogram of speed and angle of drawing vectors derived from an online signature, the speed and angle histogram of the simplest signature is one where all elements fall in the same bin. Therefore, the signature complexity can be defined in terms of the change magnitude and frequency in the angle and speed of drawing vectors. That is, we first define the basic drawing vector of the signature as the vector that appears the most in the signature, i.e., the histogram bin that has the highest count. Then the signature complexity is defined as the minimum earth moving distance~\cite{swaminathan2017new} required to convert the histogram of signature to the histogram of simplest signature as discussed above. That is:
\begin{equation}
EMD(T) = \sum_{i=1}^{M}\sum_{j=1}^{N}{h_T(i,j) \times w(i,j)}
\end{equation}
where ${h_T(i,j)}$ is the minimum value of histogram frequency in the $(i,j)$th histogram bin among the enrolled signatures.
The weight for each speed-angle histogram bin $w(i,j)$ is computed as the Euclidean distance from that speed-angle histogram bin to the speed-angle histogram bin of the basic drawing vector. That is, let the bin of basic drawing vector of the histogram be $(i_{ref},j_{ref})$, the distance between this bin and $(i,j)$ bin is defined as:
\begin{equation}
w(i,j) = \sqrt{\frac{(i_{ref}-i)^2}{M}+\frac{(j_{ref}-j)^2}{N}}
\end{equation}
where the size of speed-angle histogram is $M$ by $N$.

\subsubsection{Inverse of signature template dispersion:}
Naturally, it would require higher skill to forge a consistent signature without being detected. In other words, the more consistent signature is the one that is harder to forge. Therefore, dispersion of a signature or signature histogram features could be used as another factor to infer complexity of the online signature. Here the inverse of signature template dispersion $InvD(T)$ is defined as an average of inverse of index of dispersion of all features:
    \begin{equation}
    InvD(T) = \frac{1}{K}\sum_{i=1}^{K}{id^{-1}_{f_i}}
    \end{equation}
    where the index of dispersion of each histogram features, $id_{f_i}$, is formulated as the ratio of the variance to the mean of the $i^{th}$ feature to quantify whether a set of observed occurrences are clustered or dispersed compared to a standard statistical model. That is:
    \begin{equation}
    id_{f_i} = \frac{\sigma_{f_i}^2}{\mu_{f_i}}
    \end{equation}
    And $K$ is the total number of histogram features ${f_i}$ used to compute dispersion index that result in a definite number (removing all the features with zero mean).
%\end{enumerate}

Then, the complexity $C(T)$ of an online signature template $T$ can be derived using these two functions:

\begin{equation}
C(T) = EMD(T) \times InvD(T)
\end{equation}

Note that since the simplest form of signature is defined as the signature that has no change in the drawing angle or speed, the histogram features used to compute energy of signature template and inverse of dispersion of signature templates are the ones from speed-angle derived from the first and second halves of an online signature.

\subsection{Template repeatability}
\label{subsection:repeatability}
In general, the system should always grant access to legitimate users who present genuine credentials. However, in an online signature verification system, repeatability or the ability of users in repeating their own signatures can vary. As a result, genuine attempts of some users are rejected more often than others. 
Specifically, given a signature template, this paper proposed signature repeatability score that could be used to infer FRR of that template and it could be derived from a validation set of online signature samples that are collected in different sessions than the enrollment session.

Let $S(g_i|T)$ be the dissimilarity score of the genuine sample $g_i$ (computed from a recognition algorithm) derived from the template T. The inverse of repeatability score is computed from the average of the dissimilarity scores of all genuine samples in a validation set. Here, dissimilarity score could be computed from any recognition algorithm and it is the inverse of similarity score. (More details on the recognition algorithms used in this paper are provided in the next section.) That is, the inverse of repeatability score $R^{-1}(T)$ is given by:
\begin{equation}
R^{-1}(T) = \frac{\sum_{i=1}^{n}{S(g_i|T)}}{n}
\end{equation}
Or repeatability score $R(T)$ is:
\begin{equation}
R(T) = \frac{n}{\sum_{i=1}^{n}{S(g_i|T)}}
\end{equation}
where $G = \{g_i|i = 1,2,...,n\}$ is the validation set of $n$ genuine samples.

Here, the template with higher repeatability score is the one that corresponds to the user whose dissimilarity scores of genuine samples from a validation set are lower. In other words, it corresponds to a user whose genuine samples are more consistent across the sessions. As a result, for a given template, a higher repeatability score infers lower FRR.

\section{Experimental results}
\label{section:Experimental}
In this section, experiments performed to evaluate the efficacy of the proposed distinctiveness, complexity, and repeatability metrics are described and results are reported.
\subsection{Dataset}
In this study, online signature experiments were performed using two public datasets: MCYT~\cite{ortega2003mcyt} and SUSIG~\cite{kholmatov2009susig}, as well as NYU dataset~\cite{6786375}. Specifically, MCYT consists of 100 user's signatures with 25 samples per user. SUSIG consists of 94 user's signatures, 20 samples per user. NYU consists of 178 user's signatures, 30 samples per user over 6 separate sessions (5 samples each session). Unlike the first two dataset, signatures from NYU were collected in an uncontrolled and unsupervised manner using the user's iOS device via HTML5. In addition, CMU keystroke dynamics dataset~\cite{killourhy2009comparing} was also used to evaluate effectiveness of the repeatability metric. In this dataset, 400 genuine samples per user were collected from 8 sessions (50 samples per session). 

\subsection{Performance of the distinctiveness measure}

\subsubsection{Experimental protocol and verification algorithm}
\interfootnotelinepenalty=10000
The experiments were performed on MCYT, SUSIG and NYU datasets. The template of each signature was constructed using five genuine samples chosen at random from all genuine samples. This random process was repeated 100 times in MCYT dataset and SUSIG dataset, and 10 times in NYU dataset in order to get variants of templates from the same signature.

Two function-based and one feature-based algorithms were used. The two function-based algorithms are DTW and HMM approach and the feature based algorithm is the histogram approach~\cite{6786375}. For DTW approach, we have implemented the algorithm using 4 sequences which are x-y coordinates of an online signature and their first derivatives. For the x-y coordinate, the sequence is translated such that the first coordinate is at the origin. Then the pairwise distance between two signature is computed from the DTW distance normalized by the minimum length between the two signatures. Lastly, the dissimilarity score was the ratio between the average of all pairwise distance between enrolled signatures and a test signature and the average of all pairwise distance between two enrolled signatures. For HMM approach, we have used the score result from the Biosecure reference system\footnote{More information about this system is available at \url{http://svnext.it-sudparis.eu/svnview2-eph/ref_syst//}.}. Note that, for HMM approach, only the scores of signatures from MCYT dataset were available for download. Therefore, we report the performance of the proposed distinctiveness score on HMM approach for only MCYT dataset. For the implementation of histogram approach, we have used the code provided by the authors\footnote{Matlab code of this algorithm is available for download at \url{http://isis.poly.edu/~benapa/OnlineSignature/signature_index.html}.}.
The random forgery verification performance of these algorithms using the protocol described above was as follows. EERs for DTW on MCYT, SUSIG, and NYU, were 2.19\%, 3.11\%, and 5.78\%, respectively. Similarly, EERs for histogram-based were 0.72\%, 1.95\%, and 4.11\%, respectively. EERs for HMM on MCYT was 0.95\%.

At the same decision threshold for similarity score, FAR is expected to increase as distinctiveness score decreases. To demonstrate the effectiveness of the proposed template distinctiveness, the distinctiveness scores for all signature templates are first computed. Then, all the scores were sorted and the templates were divided into four groups according to their distinctiveness. For each group, the average FAR of all the users in the group at different decision threshold were computed where negative samples were drawn from a random sample of all other users. As such, the group of templates with lower distinctiveness score is expected to have higher average FAR across the threshold range.

\subsubsection{Experimental results}

\begin{figure*}
    \centering
    \subfloat[Samples of the 159 distinctiveness ]{\includegraphics[width=\columnwidth]{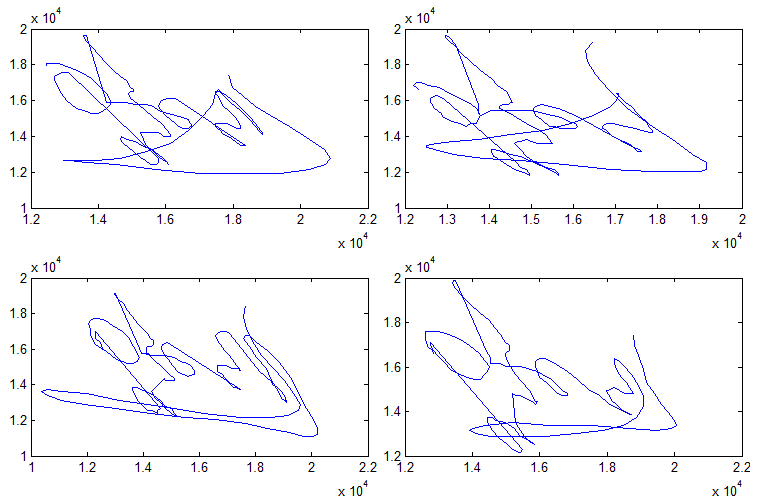}}\quad
    \subfloat[Samples of the 197 distinctiveness ]{\includegraphics[width=\columnwidth]{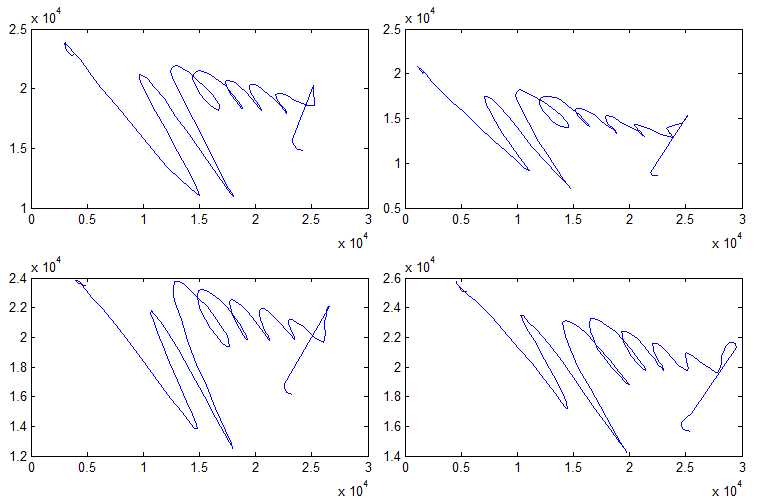}}\\
    \vspace{-0.5em}
    \subfloat[Speed-angle of the template in a)]{\includegraphics[width=\columnwidth]{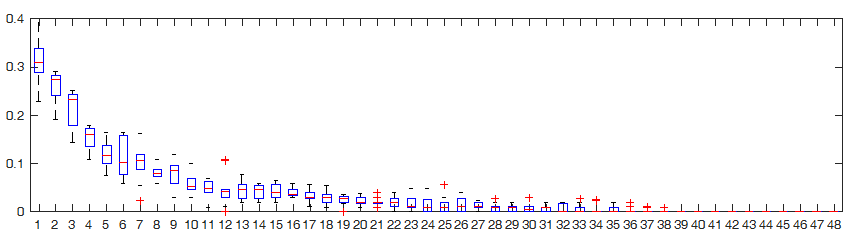}}\quad
    \subfloat[Speed-angle of the template in b)]{\includegraphics[width=\columnwidth]{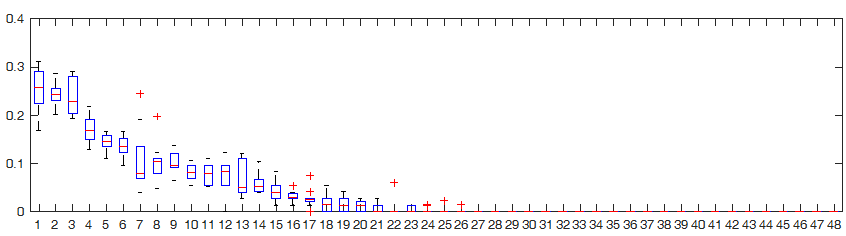}}\\
    \vspace{-0.5em}
    \caption{Samples %The 1st, 3rd, 7th, and 10th samples
    and sorted speed-angle histogram feature box-plot (by mean value) of signatures from SUSIG dataset that their templates correspond to the low and high distinctiveness% where distinctiveness is the measurement of both uniqueness and consistency of online signature samples in the enrollment
    }
    \label{fig:SUSIG_sample}
\end{figure*}

To derive the distinctiveness for each template, a signature length of 147 (the average length of signatures from MCYT and SUSIG datasets) was used to derive statistics of the population. Also, when computing the distinctiveness for a signature template in both MCYT and SUSIG, the histograms of speed-angle and pressure for the first and second halves of the signature were used since these are derived from all three attributes (x-y coordinates and pressure information) captured from a signature sample. Note that since pressure information was not available for NYU dataset, only the histograms of speed-angle for the first and second halves of the signature were used. The mean (and standard deviation) of the distinctiveness scores for templates in MCYT, SUSIG, and NYU datasets were 187.28 (23.24), 171.83 (17.73), and 108.50 (13.92), respectively.
The signature samples that resulted in the high and low distinctiveness and their feature box-plot are illustrated in Figure~\ref{fig:SUSIG_sample}. While the two signatures are somewhat similar in terms of uniqueness, samples of the template with the higher distinctiveness are more consistent than those of the other. Also, the feature box-plot shows that the signature template with the lower distinctiveness is the one with higher number of speed-angle histogram features with non-zeroes variance.

\begin{figure*}
    \centering
    \subfloat[HMM on MCYT]{\includegraphics[width=0.33\textwidth]{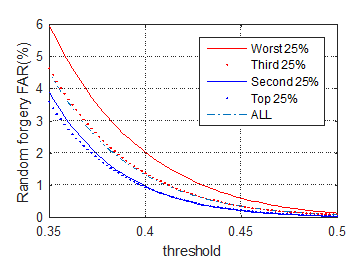}}
    \subfloat[DTW on MCYT]{\includegraphics[width=0.33\textwidth]{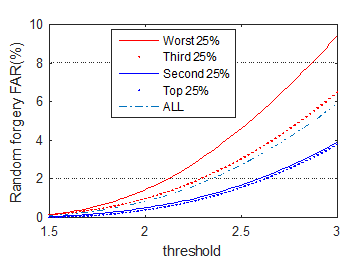}}
    \subfloat[Histogram on MCYT ]{\includegraphics[width=0.33\textwidth]{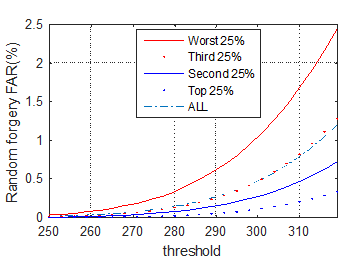}}\\
        \vspace{-0.5em}
    \subfloat[DTW on SUSIG]{\includegraphics[width=0.45\textwidth]{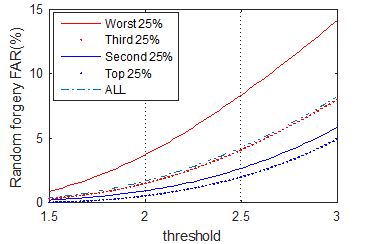}}\quad
    \subfloat[Histogram on SUSIG ]{\includegraphics[width=0.45\textwidth]{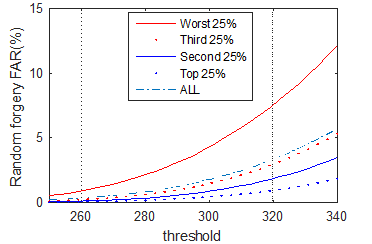}}\\
        \vspace{-0.5em}
    \subfloat[DTW on NYU]{\includegraphics[width=0.45\textwidth]{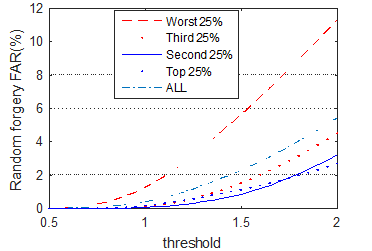}}\quad
    \subfloat[Histogram on NYU ]{\includegraphics[width=0.45\textwidth]{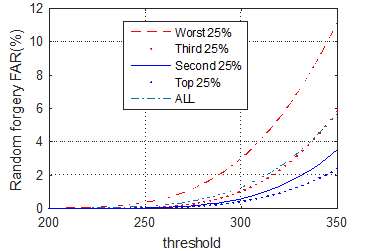}}
    \vspace{-0.5em}
    \caption{Trade-off curves between FAR of random forgery of three algorithms on MCYT, SUSIG, and NYU dataset versus threshold for each user group separated by the proposed distinctiveness}
    \label{fig:MCYT_FAR_RF}
\end{figure*}

In Figure~\ref{fig:MCYT_FAR_RF}, the dependency between FAR and template distinctiveness scores is demonstrated through trade-off curves between FAR and threshold. For MCYT and SUSIG, it is noticed that, at the same threshold level, the group with the higher distinctiveness score generally achieved better FAR. However, the decision threshold level that correspond to the same FAR was different in these two datasets. One possible explanation is that, in SUSIG dataset, the users did not use their real signatures, but instead were asked to create new signatures specifically for experimental purpose~\cite{kholmatov2009susig}. Consequently, signatures in this dataset were expected to exhibit a higher level of variation. As a result, a lower FAR was achieved at the same threshold, as compared to MCYT.

Similarly, in NYU dataset, the group of template with lower distinctiveness score consistently corresponds to higher average FAR across the threshold range. This result confirms the efficacy of the proposed distinctiveness metric regardless of dataset characteristics and the algorithms used. However, it should be note that the range of threshold and distinctiveness scores of signature templates are not directly comparable with that of those two datasets due to unavailability of pressure information in this NYU dataset. 

Lastly, we acknowledge that, since negative samples were drawn from samples of all other users in the same dataset, the trade-off curve between FAR and the threshold for each individual user was dataset dependent. Correspondingly, this rate could possibly change drastically when the same set of enrolled samples is tested against a different dataset. 

\subsubsection{Effectiveness of the random signature assumption}
In the previous subsection,
distinctiveness was derived from the generic assumption made about random signatures as described in Section~\ref{subsection:distinctiveness}. Alternately, distinctiveness could also be derived using the statistics of actual signatures (means and variances of histogram features that are used to compute distinctiveness scores). This subsection will demonstrate the effectiveness of the generic assumption made about random signatures that resulted in accurate distinctiveness scores as compared to those derived from actual signatures. First, we begin by introducing Spearman's rank correlation coefficient as a performance assessment of biometric quality metrics.
Then, this coefficient will be used to evaluate the effectiveness of the statistics derived from random signature assumption as compared to ones derived from other datasets.
Specifically, Spearman's rank coefficient is a (non-parametric) measure of statistical dependence, or strength of monotonic functional relationship, between ranks of two variables. That is:
\begin{equation}
\rho = 1 - \frac{6\times\sum_{i}^{n}{d_i^2}}{n\times (n^2-1)}\\%121.8670-susig, 215.3070--mcyt
\end{equation}
where $d_i = $rank$(x_i) - $rank$(y_i)$, rank$(x_i)$ and rank$(y_i)$ are the ranks of two variables $x_i$ and $y_i$ of the $i^{th}$ observation, and $n$ is the total number of observations. Hence, using this Spearman's rank correlation coefficient between the rank of prediction of a template's verification performance (or the rank of template's distinctiveness scores), and the rank of actual verification performance of templates (or a golden rank), the effectiveness of quality metrics can be assessed more objectively and they can be compared more easily.

One remaining task is to compute the golden rank, or ground truth of each biometric template, to be used for comparison. Since this metric will be used to evaluate distinctiveness of the template which is indicative of FAR for random forgery, this problem is addressed by computing the golden rank from the rank of an average (mean) imposter score for that user. Specifically, imposter scores are computed from  random signatures drawn from all signatures of all other users within the same dataset. As a result, the higher the Spearman's rank correlation coefficient of a quality metric means the better alignment of the rank of the quality scores with the average imposter scores of users, and, generally, the better predictive of FAR for  random forgeries. Note that, the golden rank could be adjusted to match a more specific FAR range. 

The results of Spearman's rank correlation coefficient when the statistics are derived from the generic assumption about random signatures as well as signatures from different datasets are presented in Table~\ref{table: Spearman}. 
Specifically, in this experiment, distinctiveness scores of online signature templates are derived from histogram feature statistics of each dataset in addition to the statistics derived from the generic assumption about random signatures. Then, the Spearman's rank correlation coefficient between the golden rank and the rank of distinctiveness scores derived from different statistics are computed. According to the result, the highest correlation coefficients are observed when the statistics are derived from the test dataset, except one case when DTW algorithm is applied on MCYT. For example, when computing the distinctiveness score for signature templates of MCYT dataset, the most effective statistics are the ones derived from MCYT dataset. This is not surprising as the golden rank is derived from the mean imposter scores of each template where those imposter samples are also drawn from all other signatures in the dataset.

However, in most cases, the correlation coefficient decreases greatly when statistics are derived from signatures of different datasets and these coefficients are often lower than the one when statistics are derived from generic assumption about random signatures. %This observation holds for two datasets used in the experiment, except in .
This confirms that the generic assumption can effectively be used to derive distinctiveness of a signature collection in different settings. As a result, the proposed distinctiveness metric can be computed without the need of a training set and, more importantly, without loss of generality. 
    \begin{table}
    \scriptsize
    \caption{Spearman's rank correlation coefficient between different statistics and distinctiveness scores of user's signature templates}
    \centering
%    \scalebox{1}{
    \begin{tabular}{|c|c|c|c|c|c|c|}
    \hline
    The source of &\multicolumn{3}{|c|}{DTW}&\multicolumn{3}{|c|}{Histogram}\\
    \cline{2-7}
    feature statistics&\multicolumn{3}{c|}{Test dataset}&\multicolumn{3}{c|}{Test dataset}\\
    \cline{2-7}
&MCYT&SUSIG&NYU&MCYT&SUSIG&NYU\\
    \hline
    MCYT&{0.27}&0.06&0.082&{\bf 0.78}&0.37&0.33\\%\cellcolor[gray]{0.7}
    SUSIG&0.15&{\bf 0.32}&{\bf 0.3}&0.47&{\bf 0.68}&0.59\\
    NYU&n.a.&n.a.&0.28&n.a.&n.a.&{\bf 0.71}\\
    \hline
    {Generic assumption}&{\bf 0.34}&{0.19}&{0.17}&{0.50}&{0.57}&{0.56}\\
    \hline

    \end{tabular}%}

    \label{table: Spearman}
    \end{table} 
\subsection{Performance of the complexity measure}

\subsubsection{Experimental protocol and Verification algorithm}
For complexity, experiments were performed on MCYT and SUSIG datasets since these skilled forgery samples or imitation samples are available. The random process to select 5 enrolled samples were repeated for 100 times in MCYT and SUSIG dataset, and 10 times in NYU dataset.
Since this experiment was performed on the same two datasets as the pervious one, we have used the same three verification algorithms to evaluate the effectiveness of this metric. 
The skilled forgery verification performance of these three algorithms using the protocol described above was as follows. EERs of DTW on MCYT and SUSIG were 8.28\% and 8.05, respectively. EERs of histogram-based on MCYT and SUSIG were 2.89\% and 3.93\%, respectively. EERs of HMM on MCYT was 3.41\%.

\subsubsection{Experimental results}

\begin{figure*}
    \centering
    \subfloat[HMM on MCYT]{\includegraphics[width=0.33\textwidth]{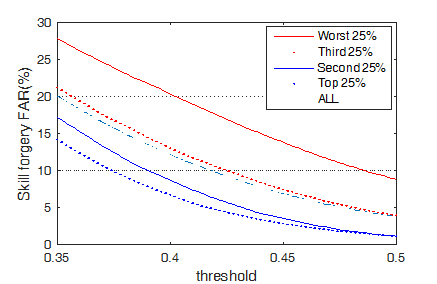}}
    \subfloat[DTW on MCYT]{\includegraphics[width=0.33\textwidth]{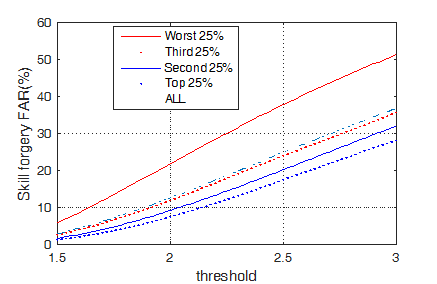}}
    \subfloat[Histogram on MCYT ]{\includegraphics[width=0.33\textwidth]{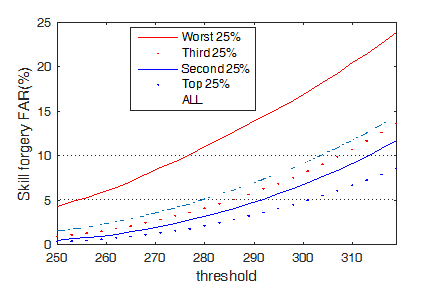}}\\
%    \vspace{-1em}
    \subfloat[DTW on SUSIG]{\includegraphics[width=0.45\textwidth]{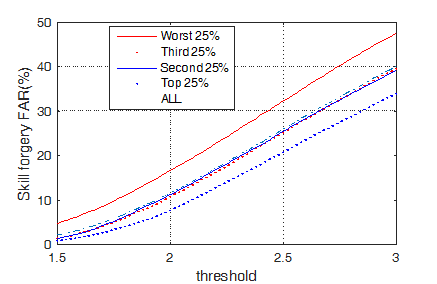}}
    \subfloat[Histogram on SUSIG]{\includegraphics[width=0.45\textwidth]{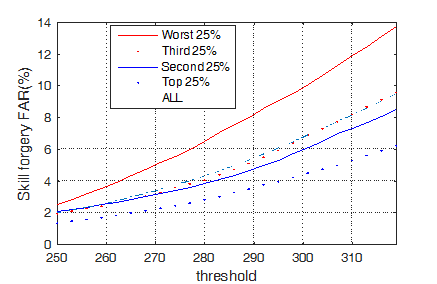}}
    \vspace{-0.5em}
    \caption{Trade-off curves between FAR of skilled forgery of three algorithms on MCYT and SUSIG dataset versus threshold for each user group separated by the proposed complexity}
    \label{fig:MCYT_FAR}
\end{figure*}

The mean (and standard deviation) of complexity scores for templates in MCYT and SUSIG datasets were 175.11 (753.11) and 145.32 (866.23), respectively.
Trade-off curves between FAR against skilled forgery for HMM, DTW, and histogram feature algorithms on MCYT and SUSIG datasets when signature templates are divided into four groups according to their complexity scores are presented in Figure~\ref{fig:MCYT_FAR}. It is noticed that, at the same threshold level, the signature group with the higher complexity score could achieve better FAR against skilled forgery. This result is consistent on both datasets and across all three verification algorithms. 

\subsubsection{Effectiveness of complexity measure}

In this subsection, the performance of the complexity measure in indicating the FAR of skilled forgeries is also analyzed in terms of the Spearman Rank correlation. That is, the Spearman Rank correlation between three different statistics of imposter dissimilarity scores and the template complexity scores as well as distinctiveness scores was computed. The three statistics of the imposter dissimilarity scores derived from skilled forgery samples were: 1) the minimum imposter score (derived from the closest forgery sample), 2) the average of the three lowest imposter scores, and 3) the average of all imposter scores. The results obtained for MCYT and SUSIG dataset are summarized in Table%~\ref{table: spearman_FARSF_MCYT} and
~\ref{table: spearman_FARSF_SUSIG}. Note that, for each signature in MCYT dataset, five forgers were chosen at random where each forger provided five imitation samples. And in SUSIG dataset, there were two types of forgeries. One was the so-called {\em"skilled forgery"} where one forger is chosen at random to provide five imitation samples for each signature after watching animation of a targeted signature on a monitor. The other was the so-called {\em"highly skilled forgery"} where two forgers (the authors of~\cite{kholmatov2009susig}) performed imitation attacks for all signatures with the knowledge of signature dynamic mapped on the screen to be traced over.

    \begin{table*}
    \small
    \setlength\tabcolsep{1.5pt}
    \caption{Spearman's rank correlation coefficient between different statistics and complexity scores (denoted as Com.) as well as distinctiveness scores (denoted as Dis.) of user's signature templates derived from MCYT and SUSIG dataset}
    \centering
%    \scalebox{1}{
    %    \scalebox{1}{
    \begin{tabular}{|c|c|c|c|c|c|c|}

\multicolumn{7}{c}{MCYT}\\
    \hline
%    \cline{2-7}
    Statistics&\multicolumn{2}{c|}{HMM}&\multicolumn{2}{c|}{DTW}&\multicolumn{2}{c|}{Histogram}\\
    \cline{2-7}
    &Com.&Dis.&Com.&Dis.&Com.&Dis.\\
    \hline
    The lowest score&0.27&0.08&0.28&0.11&0
    30&0.09\\
    Mean of 5 lowest ones&0.24&0.08&0.30&0.12&0.30&0.13\\
    Mean of all scores&0.23&0.06&0.27&0.10&0.29&0.17\\
    \hline
\multicolumn{7}{c}{}\\

    \end{tabular}%}
    \centering
%    \vspace{-1em}
    \begin{tabular}{|c|c|c|c|c|c|c|c|c|}
\multicolumn{9}{c}{SUSIG}\\
    \hline
    \cline{2-9}
    Statistics&\multicolumn{4}{c|}{DTW}&\multicolumn{4}{c|}{Histogram}\\
    \cline{2-9}
    &\multicolumn{2}{c|}{Highly Skilled}&\multicolumn{2}{c|}{Skilled}&\multicolumn{2}{c|}{Highly Skilled}&\multicolumn{2}{c|}{Skilled}\\
    \cline{2-9}
    &Com.&Dis.&Com.&Dis.&Com.&Dis.&Com.&Dis.\\
    \hline
    The lowest score&0.01&0.12&0.28&0.18&0.25&0.20&0.21&0.15\\
    Mean of 5 lowest ones&0.03&0.15&0.22&0.15&0.27&0.20&0.20&0.11\\
    Mean of all scores&0.01&0.12&0.28&0.18&0.25&0.20&0.21&0.15\\
    \hline

    \end{tabular}%}
    \label{table: spearman_FARSF_SUSIG}
    \end{table*}
    
    \begin{figure*}
    \centering
    \subfloat[NYU]{\includegraphics[width=0.4\textwidth]{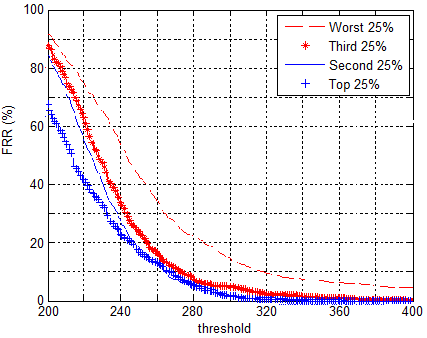}}
    \subfloat[CMU keystroke]{\includegraphics[width=0.4\textwidth]{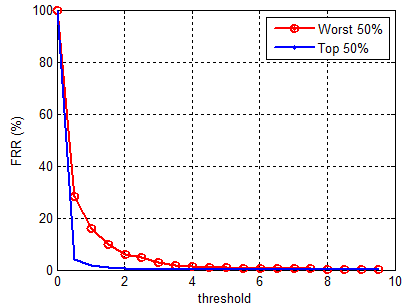}}
         \vspace{-0.5em}
    \caption{Trade-off curves between FRR versus decision threshold for each user group separated by the proposed template repeatability}
    \label{fig:mobile_FRR}
\end{figure*}

It is seen that, in cases where forgers are chosen at random, the correlation coefficient between the golden rank and the proposed complexity measure is always stronger than the ones between the golden rank and the proposed distinctiveness measure. However, in SUSIG dataset, where two forgers imitated all forgery samples (highly-skilled forgery), one can observe a much weaker correlation coefficient between the golden rank and the proposed complexity measure when DTW algorithm is applied. The difference between skilled and highly-skilled forgeries in SUSIG dataset is the expertise of the forgers and the content revealed to forgers. That is, for skilled forgery, the forgers were chosen at random and the content revealed was the animated signature on the display. And the highly-skilled forgeries were done by the two authors of the paper and the content revealed was the animated signatures on the display as well as on the screen for the forgers to trace over. This knowledge of signature structure and dynamics mapped on the screen in the highly-skilled forgery model provides an advantage to attackers in view of DTW recognition algorithm. Nevertheless, the correlation coefficient between the golden rank and the proposed complexity measure remains stronger than that of the distinctiveness measure when the histogram based approach is applied where the dynamic content of signature samples is partially destroyed while computing features.

The above results imply that analyzing the FAR of online signature templates against imitation forgery by random forgers is different from analyzing the FAR of online signature templates against random forgery. In addition, it confirms the effectiveness of the proposed complexity measure in indicating FAR against imitation forgery (the so-called skilled forgery) particularly for random forgers. However, future study to investigate performance of the proposed complexity performance against proficient forgers is still needed.
\subsection{Performance of the repeatability measure}

\subsubsection{Experimental protocol and Verification algorithm}

Since the genuine samples of MCYT and SUSIG dataset were only collected in one or two separate sessions, these two datasets are not used in this experiment. %to perform compute repeatability metric which needs measurement across sessions.
Therefore, NYU online signature dataset and CMU keystroke dynamics dataset~\cite{killourhy2009comparing} are the two datasets used to evaluate effectiveness of the repeatability metric.

For NYU datatset, the first ten samples were used to derive a user-specific online signature template and the next five were used to compute repeatability score for each user. Then, the rest fifteen ones were used to compute FRR. The verification algorithm used in this dataset is the histogram approach described in ~\cite{6786375}. Repeatability score here is computed from dissimilarity scores derived from the histogram approach. Specifically, the dissimilarity score is computed as the manhattan distance between the quantized template $F'_{template}$, or the quantized mean feature vector of the enrolled samples, and the quantized feature vector of a test signature $F'_{test}$. That is, $score = ||F'_{template}-F'_{test}|| = \sum_{i=1}^n |f'_{template}(i)-f'_{test}(i)|$  where $n$ is the number of features and $F' = \frac{F}{Q}$. And the quantization vector $Q$ is computed from standard deviation of feature vector $F$ derived from the enrolled samples. For CMU keystroke dynamics dataset, there were 400 genuine samples per user were collected from 8 sessions. The first 200 are used to train a user specific classification model using an Euclidean distance approach. The last 200 samples were divided into two parts consisting of 50 samples and 150 samples. The first 50 samples were used to compute repeatability score for each user. The second part was then used to compute FRR for each user. The algorithm used in this dataset is the feature based approach using 31 timing features, as described in~\cite{killourhy2009comparing}. The repeatability score here is computed from dissimilarity scores derived from this feature based approach. Speciﬁcally, the dissimilarity score is the squared Euclidean distance between the template $F_{template}$, or the mean feature vector of the enrolled samples, and the feature vector of a test signature $F_{test}$. That is, $score = ||(F_{template}-F_{test})||_2= \sum_{i=1}^{n}(f_{template}(i)-f_{test}(i))^2$, where $n$ is the number of features.

\subsubsection{Experimental results}

For NYU dataset, all signature templates were divided into four groups according to their repeatability scores. For each of the four groups, the average FRR of the users within the same group were computed at different threshold levels. The dependency between FRR and their repeatability metric is demonstrated in Figure~\ref{fig:mobile_FRR}(a). That is, at the same decision threshold, the corresponding FRR of user groups with higher repeatability is lower than that of users groups with lower repeatability. Note that, the inverse of repeatability index proposed in subsection~\ref{subsection:repeatability} is only applicable when signatures are collected from different sessions: enrollment, validation, and test. Therefore, for online signatures, only NYU dataset was used to perform this experiment.

For "CMU Keystroke Dynamics - Benchmark Data Set", the dependency result is presented in Figure~\ref{fig:mobile_FRR}(b). It aligns with the experiment performed on NYU dataset and confirms the efficacy of the proposed repeatability measure. Note that the users in this dataset are separated into just two groups since there are only 51 users in total. To provide a performance analysis of the repeatability measure, the Spearman Rank correlation was computed between the templates and their repeatability measure and the following three statistics of genuine dissimilarity scores derived from skilled forgery samples: 1) the maximum genuine score (derived from the worst genuine test sample), 2) the average of the five highest genuine scores, and 3) the average of all genuine scores. The Spearman Rank correlation between the golden rank computed from these three statistics and template repeatability scores are summarized in Table~\ref{table: spearman_FRR}. The strong correlation between the template repeatability measure and the three statistics used demonstrate that the proposed repeatability scores which are simply derived from the mean dissimilarity scores of genuine samples from the validation set can be used to indicate FRR of the templates effectively.

    \begin{table}
    \caption{Spearman's rank correlation coefficient between different set of statistics and repeatability scores of user's signature templates}
    \centering
%    \scalebox{1}{
    \begin{tabular}{|c|c|c|}
    \hline
    Statistics&\multicolumn{2}{c|}{Test dataset}\\
    \cline{2-3}
    &NYU&Keystroke\\
    \hline
    The highest score&0.62&0.89\\
    Mean of 5 highest scores&0.49&0.73\\
    Mean of all scores&0.55&0.80\\

    \hline
    \end{tabular}%}
    \label{table: spearman_FRR}
    \end{table} 
    
\subsection{Application}
\label{section:Applications}
In this subsection, an application of the three template characteristics namely distinctiveness, complexity and repeatability to detect problematic templates is demonstrated on SUSIG dataset. In addition, the application of distinctiveness and repeatability is demonstrated on NYU dataset, as in this dataset, skilled forgery samples were not available. The histogram approach was used for all experiments in this section as it provided the best verification performance. 
Also, the templates used were derived from all the samples in the first session (10 samples in SUSIG dataset and 5 samples in NYU dataset). Repeatability scores were computed from the first five samples in the subsequent session. Then, the rest of the samples were used as the genuine test samples. The verification performance in terms of FAR against random forgery (denoted as FAR-RF), FAR against imitation forgery or so-called skilled forgery (denoted as FAR-SF), and FRR when discarding 10\% of the templates with the lowest distinctiveness, complexity, and repeatability scores are presented in Table~\ref{table: performance_NYU}. Note that, SUSIG dataset is collected with supervision to ensure the quality of the provided signature samples. Nevertheless, the performance distinction between problematic templates (the ones with low distinctiveness, complexity and repeatability) and the rest in the dataset exists.
The results show that the proposed three template characteristics %distinctiveness, complexity and repeatability
can be used to detect problematic templates in order to %ensure the quality of enrolled templates and to
enhance overall verification performance of the system. Specifically, by discarding the 9 templates (10\% of all the templates in this dataset) with the lowest distinctiveness scores, the average FAR against random forgery improves from 3.05\% to 2.73\%, where the average FAR of the discarded templates is 6.05\%. Similarly, by discarding 10\% of templates with the lowest complexity scores, the average FAR against imitation forgery at the same decision threshold improves from 6.17\% to 5.53\%, where the average FAR of the discarded templates is 12.22\%. Also, by discarding 10\% of templates with the lowest complexity scores, the average FAR against imitation forgery at the same decision threshold improves from 2.98\% to 0.94\%, where the average FAR of those discarded templates is 22.22\%.

Note that in this dataset, the samples that are used to compute repeatability scores and the ones that are used as genuine test samples are from the same session as the data set only had two sessions. Nevertheless, as was seen in the previous section, the strong correlation between repeatability scores and FRR was observed even when genuine test samples are not from the same session but from the subsequent session.

\begin{figure*}
    \centering
    \subfloat[SUSIG - Random forgery]{\includegraphics[width=0.33\textwidth]{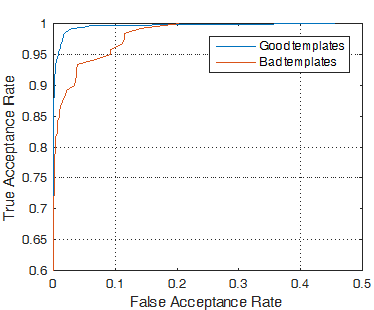}}
    \subfloat[SUSIG - Skilled forgery]{\includegraphics[width=0.3\textwidth]{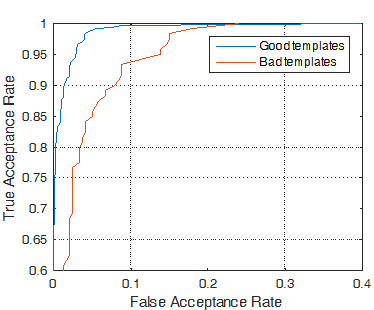}}
     \subfloat[NYU - Random forgery ]{\includegraphics[width=0.3\textwidth]{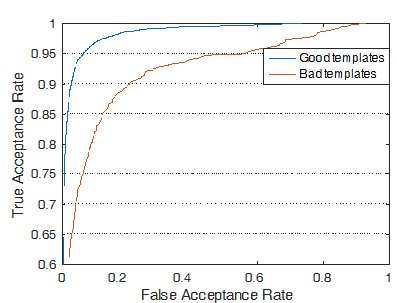}}\\
     \vspace{-0.5em}
    \caption{ ROC curves of good and bad templates in SUSIG and NYU dataset.
    }
    \label{fig:SUSIG_ROC}
\end{figure*}

Finally, when only considering templates with neither distinctiveness, or complexity or repeatability scores in the lowest 10\%, HTER (Half Total Error Rate, or the average rate between FAR and FRR) of the system decreases from 3.01\% to 2.03 (32.5\% improvement) when using the EER decision threshold against random forgery when all templates are included.  Similarly, the HTER of the system, at the same decision threshold, against imitation forgery decreases from 4.57\% to 3.21 (29.73\% improvement). In addition, the ROC comparison of good and bad templates (the templates are good when their distinctiveness, complexity, and repeatability are not in the lowest 10\%; they are bad otherwise) is depicted in Figure~\ref{fig:SUSIG_ROC}. For random forgery, the verification performance of the bad templates, is at 6.27\% EER as compared to 1.67\% EER for the good ones. For skilled forgery, the verification performance of the bad templates, is at 8.54\% EER as compared to 3.21\% EER for the good ones.

Similarly, for NYU dataset, by discarding 18 templates (10\% of all the templates in this dataset) with the lowest distinctiveness scores, the average FAR against random forgery improves from 7.51\% to 6.57\%, where the average FAR of the discarded templates is 15.85\%. Similarly, by discarding 10\% of templates with the lowest repeatability score, the average FRR improves from 7.42\% to 4.21\%, where the average FAR of the discarded templates is 36.01\%. It is observed that the discrepancy between the templates with the lowest distinctiveness and repeatability in this dataset is much higher than of those in SUSIG dataset. This could be due to the fact that, the samples in this dataset are collected without supervision, unlike SUSIG dataset. This underscores the need of the performance measures presented in this paper for mobile, unsupervised settings.
Finally, when discarding templates with either their distinctiveness or repeatability scores in the 10\% lowest score, HTER improves from 7.47\% to 5.80\% (22.37\% improvement). In addition, ROC curve comparison of good and bad templates is also depicted in Figure~\ref{fig:SUSIG_ROC} where EER of the good templates is 5.62\% as compared to 13.87\% for the bad ones.

    \begin{table}
    \caption{Verification performance of the system at EER decision threshold derived from SUSIG dataset and NYU dataset %(threshold = 390)
    when the 10\% of templates with the lowest distinctiveness, complexity, and repeatability scores are discarded}
    \centering
%    \scalebox{1}{
    \begin{tabular}{|c|c|c|c|c|}
    \multicolumn{5}{c}{SUSIG}\\
    \hline
    Constraints&$\sharp$ templates&FAR-RF&FAR-SF&FRR\\
    \hline
    None&94&3.05&6.17&2.98\\
    Distinctiveness&85&2.73&--&3.06\\
    Complexity&85&--&5.53&3.06\\
    Repeatability&85&3.17&5.88&0.94\\
    \hline
    {\bf All combined}&{\bf 70}&{\bf 2.93}&{\bf 5.29}&{\bf 1.14}\\
    \hline
    \multicolumn{5}{c}{}\\
    \end{tabular}%}

    \centering
%    \scalebox{1}{
    \begin{tabular}{|c|c|c|c|}
    \multicolumn{4}{c}{NYU}\\
    \hline
    Constraints&$\sharp$ templates&FAR-RF&FRR\\
    \hline
    None&178&7.51&7.42\\
    Distinctiveness&160&6.57&8.13\\
    Repeatability&160&7.99&4.21\\
    \hline
    {\bf All combined}&{\bf 142}&{\bf 6.99}&{\bf 4.60}\\

    \hline
    \end{tabular}%}
    \label{table: performance_NYU}
    \end{table}

These results demonstrate that the verification performance of the system against random forgery and skilled forgery could be improved simultaneously by enrolling only templates with sufficient distinctiveness, complexity, and repeatability score. More importantly, weak user's templates, in terms of these three measures could be identified in advance without leaving security and accessibility of those users in jeopardy.

The three metrics proposed in this paper could also be used to implement other strategies that may be more appropriate for a given application or for a specific user. For example, given a low distinctiveness template, the system could verify whether a few bad enrolled samples could be safely removed to enhance FAR without degrading FRR. And given a low repeatability template, the system may prompt users to practice more on their signatures. Exploring the benefits of the proposed metrics in different ways and in different applications could be a fertile area for future work.

\section{Discussion and future work}

\label{section:Discussion}
This paper presents three measures, namely distinctiveness, complexity, and repeatability, for online  signature templates. These are three important aspects for a biometric system that operates in an unsupervised setting not only because the issues are more likely to occur but when they occur they must be detected automatically. The measures are based on general statistics and can be derived from any feature-based biometric template with fixed length feature vectors and hence they can also be applied to other types of biometric recognition systems.

The proposed distinctiveness measure simultaneously incorporates the intrinsic difference as well as the consistency of a set of enrolled samples. 
The results evaluated on three datasets confirmed the efficacy of the proposed metric in distinguishing between high and low false acceptance rate templates at the same decision threshold. Moreover, Spearman Rank correlation coefficient of the distinctiveness score computed from different assumptions about random signatures demonstrated the effectiveness of the statistical assumption about random signatures as compared to alternate assumptions. In addition, the characteristics of online signature template in terms of complexity and repeatability metric were investigated. Experiments confirmed that the complexity and repeatability of templates can be inferred from an enrollment set and a validation set that is collected in sessions other than the training set, respectively. {Subsequently, performance improvements of the system when these three proposed metrics are simultaneously used to detect problematic templates was demonstrated.}

One area of future work could be to examine {other applications and training strategies} to incorporate these three measures to enhance security, usability, and reliability of an online signature verification.  In addition, these three metrics could be used to investigate characteristic differences between datasets of online signatures, and other biometric modalities, that are collected in unsupervised and supervised manner. Finally, while the performance of the proposed template characteristic assessment has been demonstrated on online signatures and keystroke dynamics, the applicability and effectiveness of these three metrics towards other biometric modalities has to be explored.

\section*{Acknowledgments}

This work was supported by Thailand Research Fund (MRG5980078), the NSF (grant 1228842), and New York University, Abu Dhabi, Cyber Security Center.
We would also like to thank Taha H. Sencar for helpful discussions.

% BALANCE COLUMNS
\balance{}

% REFERENCES FORMAT
% References must be the same font size as other body text.
\bibliographystyle{SIGCHI-Reference-Format}

%%% -*-BibTeX-*-
%%% Do NOT edit. File created by BibTeX with style
%%% ACM-Reference-Format-Journals [18-Jan-2012].

\begin{thebibliography}{00}

%%% ====================================================================
%%% NOTE TO THE USER: you can override these defaults by providing
%%% customized versions of any of these macros before the \bibliography
%%% command.  Each of them MUST provide its own final punctuation,
%%% except for \shownote{}, \showDOI{}, and \showURL{}.  The latter two
%%% do not use final punctuation, in order to avoid confusing it with
%%% the Web address.
%%%
%%% To suppress output of a particular field, define its macro to expand
%%% to an empty string, or better, \unskip, like this:
%%%
%%% \newcommand{\showDOI}[1]{\unskip}   % LaTeX syntax
%%%
%%% \def \showDOI #1{\unskip}           % plain TeX syntax
%%%
%%% ====================================================================

\ifx \showCODEN    \undefined \def \showCODEN     #1{\unskip}     \fi
\ifx \showDOI      \undefined \def \showDOI       #1{{\tt DOI:}\penalty0{#1}\ }
  \fi
\ifx \showISBNx    \undefined \def \showISBNx     #1{\unskip}     \fi
\ifx \showISBNxiii \undefined \def \showISBNxiii  #1{\unskip}     \fi
\ifx \showISSN     \undefined \def \showISSN      #1{\unskip}     \fi
\ifx \showLCCN     \undefined \def \showLCCN      #1{\unskip}     \fi
\ifx \shownote     \undefined \def \shownote      #1{#1}          \fi
\ifx \showarticletitle \undefined \def \showarticletitle #1{#1}   \fi
\ifx \showURL      \undefined \def \showURL       #1{#1}          \fi

\end{thebibliography}


\begin{thebibliography}{00}

%%% ====================================================================
%%% NOTE TO THE USER: you can override these defaults by providing
%%% customized versions of any of these macros before the \bibliography
%%% command.  Each of them MUST provide its own final punctuation,
%%% except for \shownote{}, \showDOI{}, and \showURL{}.  The latter two
%%% do not use final punctuation, in order to avoid confusing it with
%%% the Web address.
%%%
%%% To suppress output of a particular field, define its macro to expand
%%% to an empty string, or better, \unskip, like this:
%%%
%%% \newcommand{\showDOI}[1]{\unskip}   % LaTeX syntax
%%%
%%% \def \showDOI #1{\unskip}           % plain TeX syntax
%%%
%%% ====================================================================

\ifx \showCODEN    \undefined \def \showCODEN     #1{\unskip}     \fi
\ifx \showDOI      \undefined \def \showDOI       #1{{\tt DOI:}\penalty0{#1}\ }
  \fi
\ifx \showISBNx    \undefined \def \showISBNx     #1{\unskip}     \fi
\ifx \showISBNxiii \undefined \def \showISBNxiii  #1{\unskip}     \fi
\ifx \showISSN     \undefined \def \showISSN      #1{\unskip}     \fi
\ifx \showLCCN     \undefined \def \showLCCN      #1{\unskip}     \fi
\ifx \shownote     \undefined \def \shownote      #1{#1}          \fi
\ifx \showarticletitle \undefined \def \showarticletitle #1{#1}   \fi
\ifx \showURL      \undefined \def \showURL       #1{#1}          \fi

\bibitem{alonso2012quality}
{Fernando Alonso-Fernandez}, {Julian Fierrez}, {and} {Javier Ortega-Garcia}.
  2012.
\newblock \showarticletitle{Quality measures in biometric systems}.
\newblock {\em Security \& Privacy, IEEE\/} {10}, 6 (2012), 52--62.
\newblock


\bibitem{bharadwaj2015qfuse}
{Samarth Bharadwaj}, {Himanshu~S Bhatt}, {Richa Singh}, {Mayank Vatsa}, {and}
  {Afzel Noore}. 2015.
\newblock \showarticletitle{QFuse: Online learning framework for adaptive
  biometric system}.
\newblock {\em Pattern Recognition\/} {48}, 11 (2015), 3428--3439.
\newblock


\bibitem{cho2006artificial}
{Sungzoon Cho} {and} {Seongseob Hwang}. 2006.
\newblock \showarticletitle{Artificial rhythms and cues for keystroke dynamics
  based authentication}.
\newblock In {\em Advances in Biometrics}. Springer, 626--632.
\newblock


\bibitem{cpalka2014new}
{Krzysztof Cpa{\l}ka}, {Marcin Zalasi{\'n}ski}, {and} {Leszek Rutkowski}. 2014.
\newblock \showarticletitle{New method for the on-line signature verification
  based on horizontal partitioning}.
\newblock {\em Pattern Recognition\/} {47}, 8 (2014), 2652--2661.
\newblock


\bibitem{daugman2003importance}
{John Daugman}. 2003.
\newblock \showarticletitle{The importance of being random: statistical
  principles of iris recognition}.
\newblock {\em Pattern recognition\/} {36}, 2 (2003), 279--291.
\newblock


\bibitem{ferrer2017evaluation}
{Teresa Ferrer-Blasco}, {Alberto Dom{\'\i}nguez-Vicent}, {Jos{\'e}~J
  Esteve-Taboada}, {Miguel~A Aloy}, {Jose~E Adsuara}, {and} {Robert
  Mont{\'e}s-Mic{\'o}}. 2017.
\newblock \showarticletitle{Evaluation of the repeatability of a swept-source
  ocular biometer for measuring ocular biometric parameters}.
\newblock {\em Graefe's Archive for Clinical and Experimental Ophthalmology\/}
  {255}, 2 (2017), 343--349.
\newblock


\bibitem{galbally2011quality}
{Javier Galbally}, {Julian Fierrez}, {Marcos Martinez-Diaz}, {and} {R{\'e}jean
  Plamondon}. 2011.
\newblock \showarticletitle{Quality analysis of dynamic signature based on the
  sigma-lognormal model}. In {\em Document Analysis and Recognition (ICDAR),
  2011 Int'l Conf. on}. IEEE, 633--637.
\newblock


\bibitem{grother2007performance}
{Patrick Grother} {and} {Elham Tabassi}. 2007.
\newblock \showarticletitle{Performance of biometric quality measures}.
\newblock {\em Pattern Analysis and Machine Intelligence, IEEE Trans. on\/}
  {29}, 4 (2007), 531--543.
\newblock


\bibitem{guest2004repeatability}
{Richard~M Guest}. 2004.
\newblock \showarticletitle{The repeatability of signatures}. In {\em Frontiers
  in Handwriting Recognition, 2004. IWFHR-9 2004. Ninth International Workshop
  on}. IEEE, 492--497.
\newblock


\bibitem{houmani2014quality}
{Nesma Houmani} {and} {Sonia Garcia-Salicetti}. 2014.
\newblock \showarticletitle{Quality Measures for Online Handwritten
  Signatures}.
\newblock In {\em Signal and Image Processing for Biometrics}. Springer,
  255--283.
\newblock


\bibitem{houmani2016hunting}
{Nesma Houmani} {and} {Sonia Garcia-Salicetti}. 2016.
\newblock \showarticletitle{On Hunting Animals of the Biometric Menagerie for
  Online Signature}.
\newblock {\em PloS one\/} {11}, 4 (2016), e0151691.
\newblock


\bibitem{houmani2012biosecure}
{Nesma Houmani}, {Aur{\'e}lien Mayoue}, {Sonia Garcia-Salicetti}, {Bernadette
  Dorizzi}, {Mostafa~I Khalil}, {Mohamed~N Moustafa}, {Hazem Abbas}, {Daigo
  Muramatsu}, {Berrin Yanikoglu}, {Alisher Kholmatov}, {and} {others}. 2012.
\newblock \showarticletitle{{BioSecure signature evaluation campaign
  (BSEC'2009): Evaluating online signature algorithms depending on the quality
  of signatures}}.
\newblock {\em Pattern Recognition\/} {45}, 3 (2012), 993--1003.
\newblock


\bibitem{huang2015adaptive}
{Zengxi Huang}, {Yiguang Liu}, {Xuwei Li}, {and} {Jie Li}. 2015.
\newblock \showarticletitle{An adaptive bimodal recognition framework using
  sparse coding for face and ear}.
\newblock {\em Pattern Recognition Letters\/}  {53} (2015), 69--76.
\newblock


\bibitem{jain201650}
{Anil~K Jain}, {Karthik Nandakumar}, {and} {Arun Ross}. 2016.
\newblock \showarticletitle{{50 years of biometric research: Accomplishments,
  challenges, and opportunities}}.
\newblock {\em Pattern Recognition Letters\/}  {79} (2016), 80--105.
\newblock


\bibitem{kholmatov2009susig}
{Alisher Kholmatov} {and} {Berrin Yanikoglu}. 2009.
\newblock \showarticletitle{{SUSIG}: an on-line signature database, associated
  protocols and benchmark results}.
\newblock {\em Pattern Analysis and Applications\/} {12}, 3 (2009), 227--236.
\newblock


\bibitem{killourhy2009comparing}
{Kevin~S Killourhy}, {Roy Maxion}, {and} {others}. 2009.
\newblock \showarticletitle{Comparing anomaly-detection algorithms for
  keystroke dynamics}. In {\em Dependable Systems \& Networks, 2009. DSN'09.
  IEEE/IFIP International Conference on}. IEEE, 125--134.
\newblock


\bibitem{komanduri2014telepathwords}
{Saranga Komanduri}, {Richard Shay}, {Lorrie~Faith Cranor}, {Cormac Herley},
  {and} {Stuart~E Schechter}. 2014.
\newblock \showarticletitle{{Telepathwords: Preventing Weak Passwords by
  Reading Users' Minds}}. In {\em USENIX Security Symposium}. 591--606.
\newblock


\bibitem{labati2016biometric}
{Ruggero~Donida Labati}, {Angelo Genovese}, {Enrique Mu{\~n}oz}, {Vincenzo
  Piuri}, {Fabio Scotti}, {and} {Gianluca Sforza}. 2016.
\newblock \showarticletitle{{Biometric recognition in automated border control:
  a survey}}.
\newblock {\em ACM Computing Surveys (CSUR)\/} {49}, 2 (2016), 24.
\newblock


\bibitem{manjunatha2016online}
{KS Manjunatha}, {S Manjunath}, {DS Guru}, {and} {MT Somashekara}. 2016.
\newblock \showarticletitle{Online signature verification based on writer
  dependent features and classifiers}.
\newblock {\em Pattern Recognition Letters\/}  {80} (2016), 129--136.
\newblock


\bibitem{morales2014towards}
{Aythami Morales}, {Julian Fierrez}, {and} {Javier Ortega-Garcia}. 2014.
\newblock \showarticletitle{Towards predicting good users for biometric
  recognition based on keystroke dynamics}. In {\em European Conference on
  Computer Vision}. Springer, 711--724.
\newblock


\bibitem{nguyen2017smartwatches}
{Toan Nguyen} {and} {Nasir Memon}. 2017.
\newblock \showarticletitle{Smartwatches Locking Methods: A Comparative Study}.
  In {\em Symposium on Usable Privacy and Security (SOUPS)}.
\newblock


\bibitem{NGUYEN2018174}
{Toan Nguyen} {and} {Nasir Memon}. 2018.
\newblock \showarticletitle{Tap-based user authentication for smartwatches}.
\newblock {\em Computers \& Security\/}  {78} (2018), 174 -- 186.
\newblock
\showISSN{0167-4048}
\showDOI{%
\url{https://dx.doi.org/10.1016/j.cose.2018.07.001}}


\bibitem{ortega2003mcyt}
{Javier Ortega-Garcia}, {J Fierrez-Aguilar}, {D Simon}, {J Gonzalez}, {M
  Faundez-Zanuy}, {V Espinosa}, {A Satue}, {I Hernaez}, {J-J Igarza}, {C
  Vivaracho}, {and} {others}. 2003.
\newblock \showarticletitle{{MCYT} baseline corpus: a bimodal biometric
  database}.
\newblock {\em IEE Proceedings-Vision, Image and Signal Processing\/} {150}, 6
  (2003), 395--401.
\newblock


\bibitem{parodi2014legendre}
{Marianela Parodi} {and} {Juan~C G{\'o}mez}. 2014.
\newblock \showarticletitle{Legendre polynomials based feature extraction for
  online signature verification. Consistency analysis of feature combinations}.
\newblock {\em Pattern Recognition\/} {47}, 1 (2014), 128--140.
\newblock


\bibitem{poh2008incorporating}
{Norman Poh} {and} {Josef Kittler}. 2008.
\newblock \showarticletitle{Incorporating variation of model-specific score
  distribution in speaker verification systems}.
\newblock {\em IEEE Trans. on Audio, Speech and Language Processing\/} {16}, 3
  (2008), 594--606.
\newblock


\bibitem{poh2013user}
{Norman Poh}, {Arun Ross}, {Weifeng Lee}, {and} {Josef Kittler}. 2013.
\newblock \showarticletitle{A user-specific and selective multimodal biometric
  fusion strategy by ranking subjects}.
\newblock {\em Pattern Recognition\/} {46}, 12 (2013), 3341--3357.
\newblock


\bibitem{rattani2012analysis}
{Ajita Rattani}, {Norman Poh}, {and} {Arun Ross}. 2012.
\newblock \showarticletitle{Analysis of user-specific score characteristics for
  spoof biometric attacks}. In {\em Computer Vision and Pattern Recognition
  Workshops (CVPRW), 2012 IEEE Computer Society Conference on}. IEEE, 124--129.
\newblock


\bibitem{ross2009exploiting}
{Arun Ross}, {Ajita Rattani}, {and} {Massimo Tistarelli}. 2009.
\newblock \showarticletitle{Exploiting the “Doddington zoo” effect in
  biometric fusion}. In {\em Biometrics: Theory, Applications, and Systems,
  2009. BTAS'09. IEEE 3rd International Conf. on}. IEEE, 1--7.
\newblock


\bibitem{ruiz2017permanence}
{Maria~V Ruiz-Blondet}, {Zhanpeng Jin}, {and} {Sarah Laszlo}. 2017.
\newblock \showarticletitle{Permanence of the CEREBRE brain biometric
  protocol}.
\newblock {\em Pattern Recognition Letters\/}  {95} (2017), 37--43.
\newblock


\bibitem{6786375}
{Napa Sae-Bae} {and} {N. Memon}. 2014.
\newblock \showarticletitle{Online Signature Verification on Mobile Devices}.
\newblock {\em Information Forensics and Security, IEEE Trans. on\/} {9}, 6
  (June 2014), 933--947.
\newblock


\bibitem{sae2015quality}
{Napa Sae-Bae} {and} {Nasir Memon}. 2015.
\newblock \showarticletitle{Quality of online signature templates}. In {\em
  Identity, Security and Behavior Analysis (ISBA), 2015 IEEE International
  Conference on}. IEEE, 1--8.
\newblock


\bibitem{SAEBAE2018332}
{Napa Sae-Bae}, {Nasir Memon}, {and} {Pitikhate Sooraksa}. 2018.
\newblock \showarticletitle{Distinctiveness, complexity, and repeatability of
  online signature templates}.
\newblock {\em Pattern Recognition\/}  {84} (2018), 332 -- 344.
\newblock
\showISSN{0031-3203}
\showDOI{%
\url{https://dx.doi.org/10.1016/j.patcog.2018.07.024}}


\bibitem{swaminathan2017new}
{Muthukaruppan Swaminathan}, {Pankaj~Kumar Yadav}, {Obdulio Piloto}, {Tobias
  Sj{\"o}blom}, {and} {Ian Cheong}. 2017.
\newblock \showarticletitle{A new distance measure for non-identical data with
  application to image classification}.
\newblock {\em Pattern Recognition\/}  {63} (2017), 384--396.
\newblock


\bibitem{unar2014review}
{JA Unar}, {Woo~Chaw Seng}, {and} {Almas Abbasi}. 2014.
\newblock \showarticletitle{A review of biometric technology along with trends
  and prospects}.
\newblock {\em Pattern recognition\/} {47}, 8 (2014), 2673--2688.
\newblock


\bibitem{van2018user}
{Toan Van~Nguyen}. 2018.
\newblock {\em User Identification and Authentication on Emerging Interfaces}.
\newblock Ph.D. Dissertation. New York University Tandon School of Engineering.
\newblock


\bibitem{van2014finger}
{Toan Van~Nguyen}, {Napa Sae-Bae}, {and} {Nasir Memon}. 2014.
\newblock \showarticletitle{Finger-drawn pin authentication on touch devices}.
  In {\em 2014 IEEE International Conference on Image Processing (ICIP)}. IEEE,
  5002--5006.
\newblock


\bibitem{nguyen2017draw}
{Toan Van~Nguyen}, {Napa Sae-Bae}, {and} {Nasir Memon}. 2017.
\newblock \showarticletitle{DRAW-A-PIN: Authentication using finger-drawn PIN
  on touch devices}.
\newblock {\em Computers \& Security\/}  {66} (2017), 115--128.
\newblock


\bibitem{villalba2016analysis}
{Jes{\'u}s Villalba}, {Alfonso Ortega}, {Antonio Miguel}, {and} {Eduardo
  Lleida}. 2016.
\newblock \showarticletitle{Analysis of speech quality measures for the task of
  estimating the reliability of speaker verification decisions}.
\newblock {\em Speech Communication\/}  {78} (2016), 42--61.
\newblock


\bibitem{vivaracho2016client}
{Carlos Vivaracho-Pascual}, {Arancha Simon-Hurtado}, {Esperanza
  Manso-Martinez}, {and} {Juan~M Pascual-Gaspar}. 2016.
\newblock \showarticletitle{{ Client threshold prediction in biometric
  signature recognition by means of Multiple Linear Regression and its use for
  score normalization}}.
\newblock {\em Pattern Recognition\/}  {55} (2016), 1--13.
\newblock


\bibitem{yager2010biometric}
{Neil Yager} {and} {Ted Dunstone}. 2010.
\newblock \showarticletitle{The biometric menagerie}.
\newblock {\em Pattern Analysis and Machine Intelligence, IEEE Transactions
  on\/} {32}, 2 (2010), 220--230.
\newblock


\bibitem{zois2016offline}
{Elias~N Zois}, {Linda Alewijnse}, {and} {George Economou}. 2016.
\newblock \showarticletitle{{Offline signature verification and quality
  characterization using poset-oriented grid features}}.
\newblock {\em Pattern Recognition\/}  {54} (2016), 162--177.
\newblock


\end{thebibliography}

\end{document}